\pdfoutput=1

\documentclass[11pt]{article}

\usepackage[final]{acl}

\usepackage{times}
\usepackage{latexsym}

\usepackage[T1]{fontenc}

\usepackage[utf8]{inputenc}

\usepackage{microtype}

\usepackage{inconsolata}

\usepackage{graphicx}
\usepackage{amsmath}
\usepackage{amsfonts}

\usepackage{amssymb}
\usepackage{tcolorbox} 
\tcbset{
    colback=gray!10,   
    colframe=black,    
    boxrule=0.5pt,     
    arc=2mm,           
    left=2mm, right=2mm, top=1mm, bottom=1mm 
}
\title{Seeing Hate Differently: Hate Subspace Modeling for Culture-Aware Hate Speech Detection}

\author{Weibin Cai \\
  Data Lab, EECS Department \\
  Syracuse University \\
  \texttt{weibin44@data.syr.edu} \\\And
  Reza Zafarani \\
  Data Lab, EECS Department \\
  Syracuse University \\
  \texttt{reza@data.syr.edu} \\}

\definecolor{BrickRed}{rgb}{0.8, 0.25, 0.33}
\definecolor{airforceblue}{rgb}{0.36, 0.54, 0.66}
\definecolor{ballblue}{rgb}{0.13, 0.67, 0.8}

\begin{document}
\maketitle
\begin{abstract}
Hate speech detection has been extensively studied, yet existing methods often overlook a real-world complexity: training labels are biased, and interpretations of what is considered hate vary across individuals with different cultural backgrounds. We first analyze these challenges, including data sparsity, cultural entanglement, and ambiguous labeling. To address them, we propose a culture-aware framework that constructs individuals' hate subspaces. To alleviate data sparsity, we model combinations of cultural attributes. For cultural entanglement and ambiguous labels, we use label propagation to capture distinctive features of each combination. Finally, individual hate subspaces, which in turn can further enhance classification performance. Experiments show our method outperforms state-of-the-art by 1.05\% on average across all metrics.

\end{abstract}
\section{Introduction}

Hate speech detection aims to determine whether a text contains hateful content. Traditional approaches primarily focus on textual features, such as critical lexicon cues and syntactic patterns~\cite{nobata2016abusive,burnap2014hate,burnap2016us}, while recent works rely on fine-tuning pre-trained language models (PLMs)~\cite{caselli2020hatebert,koufakou2020hurtbert}, achieving strong performance with F1 scores of 0.8-0.9 on benchmark datasets.

\begin{figure}[t]
  \centering
  \includegraphics[width=0.5\textwidth]{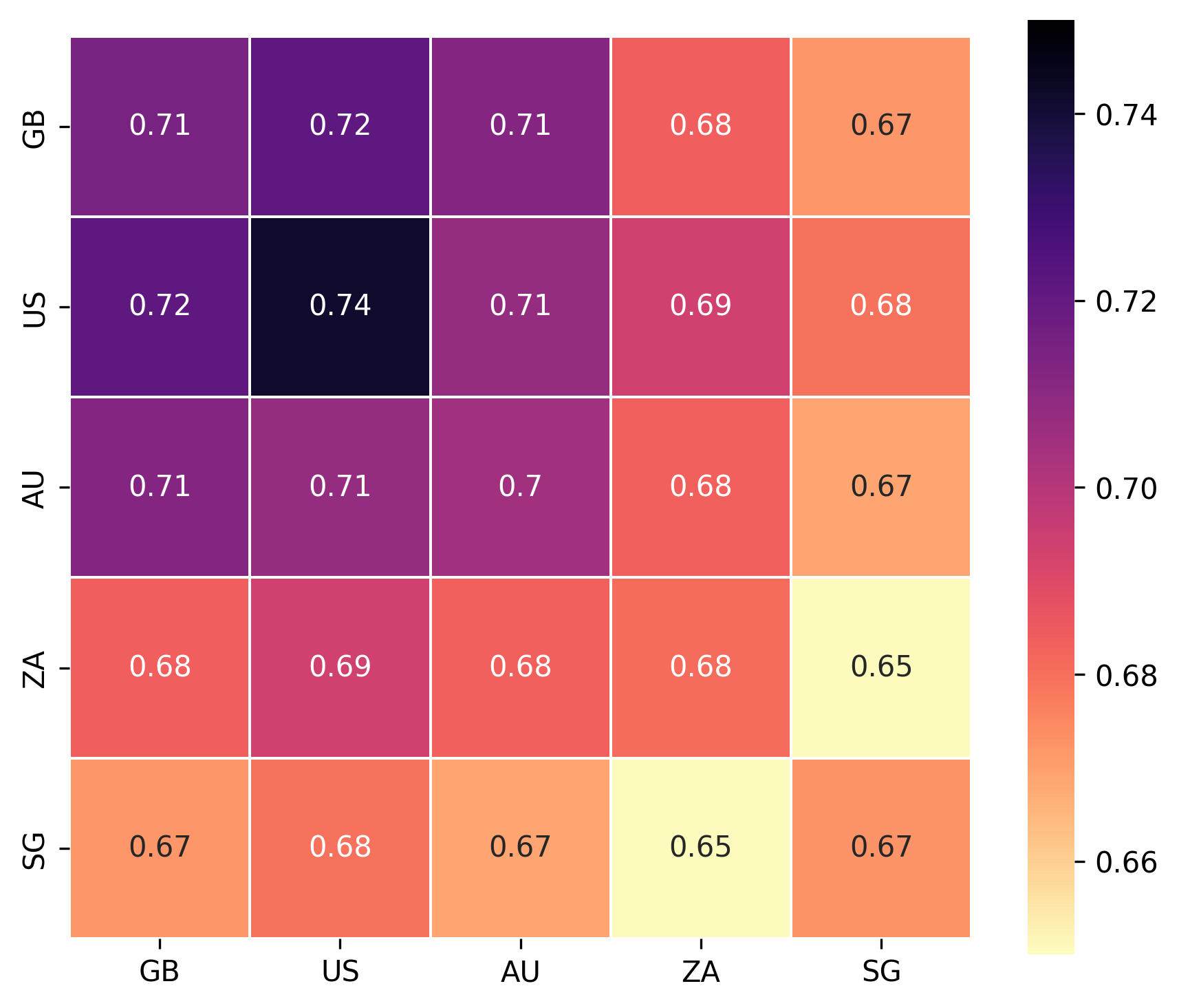}
  \caption{Pairwise hate-speech label agreement ratios between users from five countries: United Kingdom (GB), United States (US), Australia (AU), South Africa (ZA), and Singapore (SG).} 
  \label{fig:agreement}  \vspace{-6mm}
\end{figure}

However, these results can be misleading as ground-truth labels are often obtained through a majority voting among some annotators, which introduces bias and oversimplifies the problem~\cite{sap2019social}. In reality, \textbf{individuals from different cultural background may perceive the same text differently}. Prior work has shown cross-cultural disagreement in hate speech annotation~\cite{lee2023exploring}; for example, annotators from the United States and United Kingdom exhibit higher agreement than those from the United States and Singapore (see Figure \ref{fig:agreement}). Even within the same cultural group, perceptions can diverge considerably. As illustrated in Figure~\ref{fig:agreement}, pairwise label agreement ratios reveal that even annotators from the same country can have lower agreement scores compared to those of with other countries (e.g., \texttt{SG-SG} < \texttt{SG-US}). These findings suggest that hate perception is too complex to be explained by a single cultural factor. To better understand hate perception and build personalized hate speech detection systems, it is essential to uncover the diverse factors shaping individual's hate perception.

Modeling culture-aware hate speech detection presents three major challenges: \textcircled{1}\textbf{Data sparsity.} Hate perception is shaped by numerous factors——such as religion and gender——resulting in an exponential number of possible cultural background combinations. For instance, the CREHate dataset~\cite{lee2023exploring} provides eight background attributes. Excluding the continuous feature `age', the remaining seven categorical backgrounds yield 91,045,500 possible combinations (in the ideal scenario), whereas the dataset contains annotations from only 1,064 annotators, representing merely $1.17\times10^{-5}$ of the theoretical population. \textcircled{2}\textbf{Complex and abstract cultural entanglement.} It is difficult to measure how judgments shift when attributes are added or altered. For instance, suppose an annotator with background \texttt{<Country=United States, Religion=Christian>} considers a post hateful, how does this perception change if we add \texttt{<Sex=Male>}, or replace \texttt{<Religion=Christian>} with \texttt{<Religion=Buddhism>}? Current models lack the ability to capture such nuanced entanglements, a limitation also reflected in our experiments showing that LLMs fail to effectively leverage background information (Table~\ref{tab:llminf}). \textcircled{3}\textbf{Ambiguous labeling.} Cultural attributes in datasets are often incomplete, introducing label noise. Even when labels are available, it remains unclear \textit{which} cultural factors contribute to a particular judgment. For example, when an annotator with background \texttt{ <Country=United States, Religion=Christian, Sex=Male>} labels a post as hateful, the perception may stem from nationality, religion, or their joint effect. Importantly, even the annotator themselves may not be able to disentangle these factors explicitly.

In this work, we propose a culture-aware hate speech detection framework that models individual's \textit{hate subspaces} to address these challenges. To alleviate data sparsity, we model each cultural background combination in the dataset rather than individual factors. To capture the influence between cultural factors, we introduce a one-way label propagation mechanism from a cultural background combination to its subsets. Although label ambiguity remains difficult to fully resolve, we mitigate its effect by aggregating labels from higher-level combinations and constructing a weight matrix to differentiate between them. Finally, we represent each individual's hate perception using the combinations of their cultural backgrounds:
\begin{itemize}
    \item We identify key challenges in culture-aware hate speech detection and propose a simple yet effective framework that models individuals' hate subspaces based on interactions between cultural backgrounds and posts, rather than relying solely on textual features.
    \item Extensive experiments demonstrate that our approach consistently outperform state-of-the-art baselines, achieving an average improvement of 1.05\% across all metrics.
\end{itemize}

\section{Problem Statement}

\begin{table}[t]
\caption{LLM’s ability in culture-aware hate speech detection. Details are provided in Appendix~\ref{sec:prompts_zeroshot}}
\label{tab:llminf}
\centering
\resizebox{\columnwidth}{!}{
\begin{tabular}{lcccc}
\hline
\multicolumn{1}{l|}{Model}           & Accuracy   & Precision            & Recall & F1                      \\ \hline \hline
\multicolumn{1}{l|}{One for all} & 57.17   & 56.05   & 55.51          & 55.15        \\
\multicolumn{1}{l|}{+background}  & 56.79   & 55.94   & 52.89          & 47.80          \\  \hline
\multicolumn{1}{l|}{+historical labeling}    & \textbf{61.47}   & \textbf{61.49}   & \textbf{58.97}          & \textbf{57.83} \\
\multicolumn{1}{l|}{+both}  & \textit{60.31}   & \textit{59.75}   & \textit{58.09}      & \textit{57.26}          \\ \hline \vspace{-10mm}
\end{tabular}
}
\end{table} 




\textbf{Definition. Culture-Aware Hate Speech Detection.} Let $\mathcal{U}=\{(u_{1}, \mathbf{c}_1), (u_{2}, \mathbf{c}_2),\dots, (u_{n}, \mathbf{c}_n)\}$ be a set of $n$ users, where each user $u_i$ is associated with $k$ cultural background attributes (e.g., nationality, religion), denoted as $\mathbf{c}_i = \{c_{i1},c_{i2},\dots,c_{ik}\}$. Given a post $p$, the goal of culture-aware hate speech detection is to predict $P(\text{hate}|u_i,p)$ or equivalently $P(\text{hate}|\mathbf{c}_i,p)$, i.e., the likelihood that user $u_i$ would consider post $p$ as hateful, conditioned on the user's cultural background $\mathbf{c}_i$.

\section{Method}
We begin by modeling cultural hate perception (Section~\ref{sec:interaction}), which is then used to construct individual hate perception (Section~\ref{eq:individual_hate}), and is finally employed for classification (Section~\ref{sec:classification}).

\subsection{Culture-Post Interaction Matrix}
\label{sec:interaction}
To better model individual hate perception and alleviate data sparsity, we shift our modeling target from individual to specific cultural background combinations. Let $\mathcal{P}(c_i)$ denote the power set of user $u_i$'s cultural background $\mathbf{c}_i$, i.e., $\mathcal{P}(\mathbf{c}_i) = \{\, s \mid s \subseteq \{c_{i1}, c_{i2}, \dots, c_{ik}\} \,\}$ with $|\mathcal{P}(c_i)|=2^k$. For example, if $c_i=(\text{Male, US, Christian})$ then 
\[
\mathcal{P}(c_i) =
\left\{
\begin{aligned}
&\text{Male}, \text{US}, \text{Christian}, \{\text{Male, US}\} \\
&\{\text{US, Christian}\}, \{\text{Male, Christian}\}, \\
&\{\text{Male, US, Christian}\}
\end{aligned}
\right\};
\]
Our goal is to predict $P(\text{hate}|g(\mathcal{P}(\mathbf{c}_i)), p_j)$, where $g(\cdot)$ aggregates the effects of all combinations to model an individual's hate perception. Modeling combinations offers a practical benefit: even when an unseen user $u_i$ has incomplete or unseen background information, we can approximate their prediction by utilizing overlapping combinations observed in the dataset, i.e.
\begin{equation}
    P(\text{hate}|g(\mathcal{P}(\mathbf{c}_l) \cap (\bigcup_{i=1}^n \mathcal{P}(\mathbf{c}_i)), p_j).
\end{equation}
\label{eq:approximate}

To further alleviate data sparsity and label ambiguity, we aggregate annotation signals at combination-level. Concretely, for each background combination $comb_l$ and post $p_j$ we collect
\begin{equation}
\begin{split}
U_{l,j} = \{\, (u, \mathbf{c}) \in \mathcal{U} \mid\; 
& comb_l \in \mathcal{P}(\mathbf{c}), \\
& u \in Label(p_j) \,\}
\end{split}
\end{equation}
\label{eq:aggregate}
i.e., the set of users who possess combination $comb_l$ and labeled $p_j$. This aggregation implements a single-direction label propagation from observed (higher-order) combinations toward their constituent combinations: labels provided at a richer (upper) combination inform estimates for its subsets. The intuition is that while a single annotator's label does not reveal which attribute cased the judgment, pooling labels across users who share a combination yields more robust estimates of combination-level tendencies.




To further distinguish combinations, we treat each combination as a ``document'', and its contributing $(u, \mathbf{c})$ pairs as ``words''. Under this view we build a culture-post interaction matrix $Y\in\mathbb{R}^{z\times m}$ (with $z$ total combinations and $m$ posts) using TF–IDF weighting derived from aggregated labels and co-occurrence information.

\subsection{Individual Hate Perception Embedding}
\label{sec:subspace}
We factorize $Y$ to derive latent features of combinations and posts. Specifically, we initialize an embedding matrix for combinations $P \in \mathbb{R}^{z \times d}$, for posts $Q \in \mathbb{R}^{m \times d}$, and bias terms $B_c \in \mathbb{R}^{z}$ and $B_w \in \mathbb{R}^{m}$, where $d$ is the embedding dimension. The predicted score $\hat{Y}_{l, j}$ is estimated as:
\begin{equation}
    \hat{Y}_{l, j} = \mu + b_{c_{l}} + b_{w_j} + q_j^{T} p_l
    \label{eq:estimate}
\end{equation}
where $\mu$ is the global mean, $b_{c_l}$ and $b_{w_j}$ are biases, and $p_l$, $q_j$ are latent vectors of combinations and posts. To learn these embeddings, we minimize the following objective:
\begin{equation}
    \sum_{l, j} (Y_{l, j} - \hat{Y}_{l, j})^2 + \lambda(b^2_{c_l} + b^2_{w_j} + \lVert q_j \rVert^2 + \lVert p_l \rVert^2)
    \label{eq:optimize}
\end{equation}
where $\lambda$ controls regularization and prevents overfitting by penalizing large parameter values.

Individuals interpret hate speech from multiple perspectives, depending on which subset of their cultural attributes is active. Since embeddings of all combinations are aligned in the same latent space, we represent an individual's hate perception embedding as a linear combination of all its combinations:
\begin{equation}
    HP(u_i) = \sum_{comb_l \in \mathcal{P}(\mathbf{c_i})} \alpha_l \begin{bmatrix} p_l \\ b_{c_l} \end{bmatrix}
\end{equation}
\label{eq:individual_hate}
where $\alpha_l$ is a learnable coefficient reflecting the relative influence of combination $comb_l$ on $u_i$.


\subsection{Classification}
\label{sec:classification}
We integrate the individual hate perception embedding with post features for classification. Given an individual $u_i$ and a post $p_j$, the prediction is:
\begin{equation}
    P(\text{hate}|u_i,p_j) = f_{\theta}(HP(u_i), q_j, s_j)
\end{equation}
\label{eq:classification}
where $q_j$ is the post’s interaction feature from Eq.~\ref{eq:estimate}, $s_j$ is the text embedding extracted by the CLIP text encoder~\cite{radford2021learning}, and $f_\theta(\cdot)$ is a classifier with parameters $\theta$. The model predicts that $u_i$ perceives $p_j$ as hateful if $P(hate \mid u_i, p_j) \geq 0.5$, and non-hateful otherwise.

\section{Experiments}

\textbf{Dataset.} 
we conduct experiments on CREHate dataset~\cite{lee2023hate}, where each annotator has 8 different backgrounds. We randomly split the data at the post level into training/validation/test with a ratio of 70\%/15\%/15\%.

\noindent \textbf{Baselines.} To enable a  comprehensive comparison, we evaluate against two groups of baselines: (1) Pretrained language models (PLMs)~\cite{devlin2019bert, nguyen2020bertweet,caselli2020hatebert,zhang2023twhin, barbieri2020tweeteval,zhou2020challenges}: To adopt these models to our setting, we introduce additional learnable background tokens (e.g., ``[male]'' to indicate the user's gender) and prepend them to the post text before fine-tuning, following prior work~\cite{lee2023exploring}. (2) Zero-Shot Prompting: We further test LLMs in zero-shot setting, including \textit{LLama-2-7b-chat-hf} and GPT-5. The specific prompt templates and more details are described in Appendix~\ref{sec:prompts_zeroshot}~\ref{sec:culturally_adapted}.


\subsection{Classification Evaluation}
We evaluate whether models can effectively capture the relationship between text and cultural backgrounds. As shown in Table~\ref{tab:baselines}, our proposed method outperform the best baseline by an average margin of 1.05\% across all metrics. Since every model has access to the same set of posts during training, their differences in text encoding capability are minimal, which explains why PLMs achieve comparable results. This also highlights that PLMs share similar limitations in culturally-aware modeling, at least under the standard fine-tuning paradigm. Besides, although GPT-5 in the zero-shot setting yields relatively strong performance, it still lags behind fine-tuned models by a substantial margin.

\begin{table}[t]
\caption{Classification Performance}
\label{tab:baselines}
\centering
\resizebox{\columnwidth}{!}{
\begin{tabular}{lcccc}
\hline
Model         & Accuracy   & Precision            & Recall & F1                      \\ \hline \hline
HateBERT	&76.23{\scriptsize{$\pm 0.15$}}	&75.97{\scriptsize{$\pm 0.15$}}	&76.10{\scriptsize{$\pm 0.24$}}	&76.01{\scriptsize{$\pm 0.18$}} \\
Twin-BERT	&76.26{\scriptsize{$\pm 0.27$}}	&75.98{\scriptsize{$\pm 0.27$}}	&76.04{\scriptsize{$\pm 0.32$}}	&76.00{\scriptsize{$\pm 0.29$}} \\
Twitter-Roberta	&76.33{\scriptsize{$\pm 0.14$}}	&76.05{\scriptsize{$\pm 0.15$}}	&76.10{\scriptsize{$\pm 0.15$}}	&76.06{\scriptsize{$\pm 0.14$}} \\
ToDect-Roberta	&75.90{\scriptsize{$\pm 0.26$}}	&75.61{\scriptsize{$\pm 0.26$}}	&75.65{\scriptsize{$\pm 0.23$}}	&75.63{\scriptsize{$\pm 0.24$}} \\
BERT	&\textit{76.38}{\scriptsize{$\pm 0.28$}}	&\textit{76.11}{\scriptsize{$\pm 0.28$}}	&\textit{76.20}{\scriptsize{$\pm 0.35$}}	&\textit{76.14}{\scriptsize{$\pm 0.31$}} \\
BERTweet	&76.15{\scriptsize{$\pm 0.14$}}	&75.89{\scriptsize{$\pm 0.14$}}	&76.05{\scriptsize{$\pm 0.14$}}	&75.95{\scriptsize{$\pm 0.14$}} \\ \hline
LLama-2-7b-chat-hf & 56.79   & 55.94   & 52.89          & 47.80\\
GPT-5 &71.08 &70.72 &70.56 &70.63 \\ \hline

Ours &\textbf{77.37}{\scriptsize{$\pm 0.14$}} &\textbf{77.14}{\scriptsize{$\pm 0.15$}} &\textbf{77.33}{\scriptsize{$\pm 0.23$}} & \textbf{77.19}{\scriptsize{$\pm 0.17$}} \\ \hline

\end{tabular}
}
\end{table}

\begin{table}[t]
\caption{Ablation Study}
\label{tab:ablation}
\centering
\resizebox{\columnwidth}{!}{
\begin{tabular}{lcccc}
\hline
Model         & Accuracy   & Precision            & Recall & F1                      \\ \hline \hline
Ours	&\textbf{77.37}{\scriptsize{$\pm 0.14$}} &\textbf{77.14}{\scriptsize{$\pm 0.15$}} &\textbf{77.33}{\scriptsize{$\pm 0.23$}} & \textbf{77.19}{\scriptsize{$\pm 0.17$}} \\
Ours (sum)	&76.06{\scriptsize{$\pm 0.31$}}	&75.95{\scriptsize{$\pm 0.19$}}	&76.18{\scriptsize{$\pm 0.26$}}	&75.92{\scriptsize{$\pm 0.28$}} \\
Ours (mean)	&76.25{\scriptsize{$\pm 0.34$}}	&76.04{\scriptsize{$\pm 0.28$}}	&76.23{\scriptsize{$\pm 0.21$}}	&76.07{\scriptsize{$\pm 0.29$}} \\
Ours (anno)	&76.37{\scriptsize{$\pm 0.17$}}	&76.17{\scriptsize{$\pm 0.12$}}	&76.40{\scriptsize{$\pm 0.09$}}	&76.21{\scriptsize{$\pm 0.13$}} \\
\hline
$-HP(u_i)$	&76.20{\scriptsize{$\pm 0.22$}}	&76.00{\scriptsize{$\pm 0.12$}}	&76.17{\scriptsize{$\pm 0.11$}}	&76.01{\scriptsize{$\pm 0.15$}} \\ 
$-q_j$	&\textit{76.84}{\scriptsize{$\pm 0.23$}}	&\textit{76.64}{\scriptsize{$\pm 0.23$}}	&\textit{76.88}{\scriptsize{$\pm 0.24$}}	&\textit{76.69}{\scriptsize{$\pm 0.22$}} \\
$-s_j$	&76.33{\scriptsize{$\pm 0.44$}}	&76.09{\scriptsize{$\pm 0.40$}}	&76.05{\scriptsize{$\pm 0.25$}}	&76.03{\scriptsize{$\pm 0.36$}} \\
\hline 

\end{tabular}
}
\end{table}
\subsection{Ablation Study}
To validate the effectiveness of our approach, we design two sets of ablation experiments: (1) Construction of hate subspace. we compare different strategies for building individual hate perception. Specifically, we replace the weighted sum in Eq~\ref{eq:individual_hate} with sum and mean pooling, denoted as \textit{Ours (sum)} and \textit{Ours (mean)}, respectively. In addition, we replace cultural combinations with direct annotators, denoted as \textit{Ours (anno)}. The result demonstrate that modeling cultural combination, and constructing individuals' hate subspaces are helpful. (2) Contribution of each component in Eq~\ref{eq:classification}. We further assess the importance of each input feature by removing one component at a time (denoted by ``-''). The results show that an individual's hate subspace is the most critical factor. All components contribute positively to the overall performance.
\begin{figure}[t]
  \centering
  \includegraphics[width=0.5\textwidth]{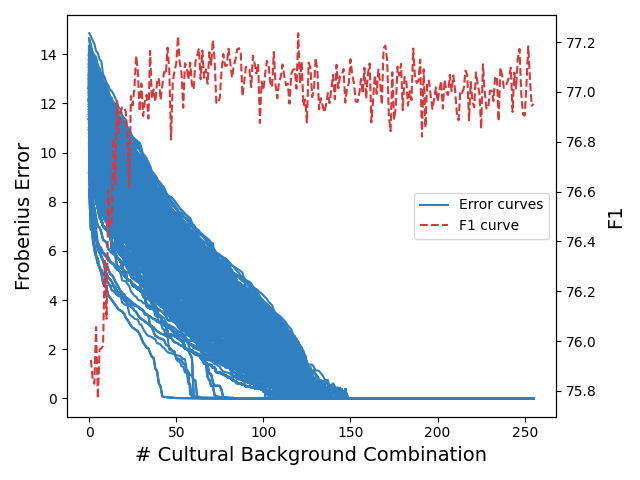}
  \caption{Effect of Number of Cultural Combinations on Hate Subspace and Classification}  
  \label{fig:error_vs_f1}  \vspace{-5mm}
\end{figure}  

\subsection{The Analysis of Hate Subspace}
In this paper, we assume that an individual's hate subspace is constructed by all possible cultural background combinations. However, the size of these combinations can be extremely large. For instance, in CREHate dataset, although there are only 1,064 annotators, the induced cultural combinations exceed 60,000. This raises both computational and modeling concerns. To analyze the complexity and effectiveness of combinations, we first compute the leverage scores of each combination within an individual's hate subspace. We then calculate the Frobenius reconstruction error when progressively adding combinations in descending order of leverage scores. As shown in Figure~\ref{fig:error_vs_f1}, fewer than half of the combinations are sufficient to reconstruct most of the hate subspaces with minimal error. We further evaluate classification performance by gradually accumulating combinations. The red curve reveals a consistent trend: performance saturates after incorporating roughly 50 combinations, and adding more combinations may introduce noise. This observation suggests that a large number of combinations are redundant and highlights an important direction for future work——understanding which combinations truly matter and how to effectively select them when constructing the hate subspace.

\section{Conclusion}
In this paper, we analyzed the challenges in culture-aware hate speech detection, focusing on three main challenges: data sparsity, the complex interplay between cultural backgrounds, and ambiguous labeling. To address these challenges, we model the interaction between cultural background combinations and posts. Specifically, we construct a culture-post interaction matrix using label propagation and apply matrix factorization to derive the hate perception of each combination. These perceptions are further used to form individuals' hate subspace, which can be leveraged to enhance classification performance. Finally, we examine the effect of the number of cultural combinations on classification. Extensive experiments demonstrate the effectiveness of our approach, yielding an average improvement of 1.05\% across all metrics.

\section{Limitation}
\label{sec:limitation}
In this paper, we propose a culture-aware model; however, it remains uncertain whether the model truly captures cultural information, and evaluating this is challenging. To investigate this, we designed an experiment using the template ``These disgusting [object]'', where ``[object]'' is replaced with a specific group, such as female or male. If the model is genuinely culture-aware, inserting a target group into ``[object]'' should lead to higher sensitivity from that group toward others, as reflected in the classifier's output.

Using the $-q_j$ version of the model, we obtain predicted scores for the female and male groups. When ``[object]'' was set to female, the mean hateful score is 0.6142 for the female group and 0.5794 for the male group. When ``[object]'' was set to male, the female group scored 0.6256 and the male group 0.6033. In both cases, the female group's scores are higher than the male group's, regardless of the target group. This may suggest that the approach fails to aware culture. However, an alternative explanation is that females may generally exhibit higher sensitivity than males. Another factor contributing to these similar scores may be the dataset itself; as shown in  Figure~\ref{fig:agreement}, distinguishing groups from a single cultural background is challenging. Therefore, while evaluating cultural awareness is important, it remains difficult to conduct reliably.

\section{Ethical Considerations}
The dataset we use is publicly available and anonymous. We do not annotate any data on our own. All the models used in this paper are publicly accessible. Their usage are consistent with their intended use. The usage of our proposed work should be used for social good. 

Although our model improves cross-cultural hate speech detection, it may still misclassify text, potentially causing unintended harm. Besides, since this work focuses on culture-aware modeling, there is a potential risk that the model could be misused to generate hate speech targeting specific cultural or demographic groups. Future work should address fairness and societal implications. The inference and finetuning of models are performed
on Quadro RTX 6000.

\section{Use of AI Assistants}
We acknowledge the use of AI language models, such as ChatGPT, for assistance in improving writing clarity and grammar. All research content is solely authored by the human authors.

\bibliography{custom}

\appendix

\section{Appendix}
\label{sec:appendix}

\subsection{Prompts for Zero-shot Experiments}
\label{sec:prompts_zeroshot}
In Table~\ref{tab:llminf}, we use LLaMA-2-7b-chat-hf to perform zero-shot experiments to test LLM's reasoning ability in personalized hate speech detection. \textit{One for all}: post content only, prediction shared by all annotators. \textit{+background}: adds annotator's cultural background to the prompt. \textit{+historical labeling}: adds 5 previously labeled posts of the annotator. \textit{+both}: includes both cultural background and historical labeling. Details are provided in Appendix~\ref{sec:prompts_zeroshot} Here are prompts for different prompts:
\begin{tcolorbox}
\small
Prompt for ``One for all'': \\

Hate speech is defined as language that attacks or demeans a person or group based on attributes like race, religion, gender, etc. Based on the given definition, determine whether people are likely to consider the following text as hate speech: POST: [post]. Is this text hateful or not? You have to answer with only one word, ``Yes'' or ``No'', do not provide any explanation or irrelevant content.
\end{tcolorbox}

\begin{tcolorbox}
\small
Prompt for ``+background'': \\

Hate speech is defined as language that attacks or demeans a person or group based on attributes like race, religion, gender, etc. Personal background: [all backgrounds]. Based on the given definition, determine whether people, based on the provided personal background information, are likely to consider the following text as hate speech: POST: [post]. Is this text hateful or not? You have to answer with only one word, ``Yes'' or ``No'', do not provide any explanation or irrelevant content. Answer:
\end{tcolorbox}

\begin{tcolorbox}
\small
Prompt for ``+historical labeling'': \\

Hate speech is defined as language that attacks or demeans a person or group based on attributes like race, religion, gender, etc. Historical labeling: [5 historical posts and corresponding labels]. Based on the given definition, determine whether people, based on their historical labeling of hate speech, are likely to consider the following text as hate speech: Post: [post]. Is this text hateful or not? You have to answer with only one word, ``Yes'' or ``No'', do not provide any explanation or irrelevant content. Answer:
\end{tcolorbox}

\begin{tcolorbox}
\small
Prompt for ``+both'': \\

Hate speech is defined as language that attacks or demeans a person or group based on attributes like race, religion, gender, etc. Personal background: [all backgrounds]. Historical labeling: [5 historical posts and corresponding labels]. Based on the given definition, determine whether people, based on their background and historical labeling of hate speech, are likely to consider the following text as hate speech: Post: [post]. Is this text hateful or not? You have to answer with only one word, ``Yes'' or ``No'', do not provide any explanation or irrelevant content. Answer:
\end{tcolorbox}

For latest LLMs, We  set instructions as ``Perform personalized hate speech classification.'' to let LLMs aware its role during inferences. Then we first give a definition of hate speech follow with a description for the task and background information: 
\begin{tcolorbox}
\small
Definition of Hate Speech: Hate speech refers to offensive discourse targeting a group or an individual based on inherent characteristics such as race, religion, sexual orientation, gender, or any other factors that may threaten social peace. \\

Answer if this post is hate or not (for people with following backgrounds:[nationality], [age], [education], [ethnicity], [gender], [politic], [religion], [gender\_sexual\_orientation] with a single alphabet letter among given answer choices a and b. \\

        POST: [post] \\
        a: Hate \\
        b: Non-hate \\
        answer: 
\end{tcolorbox}

\subsection{Culturally-adapted PLMs}
\label{sec:culturally_adapted}
Analogous to the [CLS] token in BERT, we prepend each post with trainable culture-specific tokens that serves as the representation of the corresponding cultural context. Specifically, posts associated with a given nationality are prefixed with a  [nationality] token (e.g., [Singapore]). In our scenario, since we consider multiple backgrounds, we concatenate every tokens in front of the post text, such as, ``[864] [Singapore] [100] [2] [Asian] [male] [Moderate\_liberal] [Buddhism] [heterosexual]'' follow with post text, where ``[864]'' denotes annotator id. 

\subsection{Hyperparameters}
\label{sec:parameters}
We experimented with several hyperparameter settings for fine-tuning PLMs and selected the optimal configuration: learning rate $lr=5\mathrm{e}{-6}$ and $\epsilon=1\mathrm{e}{-8}$. All experiments were run five times, and we report the mean and standard deviation. The batch size was set to 32 for all experiments. we set $lr=0.01, \lambda=0.01$ in Eq~\ref{eq:optimize}, and $d=128$.

\end{document}